\documentclass[runningheads]{llncs}
\usepackage{graphicx}

\begin{document}
\title{Conceptual Modeling and Artificial Intelligence: Mutual Benefits from Complementary Worlds}

\subtitle{Editorial preface to the 3rd Int. Workshop on Conceptual Modeling Meets Artificial Intelligence (CMAI'2021)}
\titlerunning{CM and AI: Mutual Benefits from Complementary Worlds}
\author{Dominik Bork~\orcidID{0000-0001-8259-2297}}
\authorrunning{D. Bork}
\institute{TU Wien, Business Informatics Group, Vienna, Austria\\ 
\email{dominik.bork@tuwien.ac.at}}
\maketitle              % typeset the header of the contribution
\begin{abstract}
Conceptual modeling (CM) applies abstraction to reduce the complexity of a system under study (e.g., an excerpt of reality). As a result of the conceptual modeling process a human interpretable, formalized representation (i.e., a conceptual model) is derived which enables understanding and communication among humans, and processing by machines. Artificial Intelligence (AI) algorithms are also applied to complex realities (regularly represented by vast amounts of data) to identify patterns or to classify entities in the data. Aside from the commonalities of both approaches, a significant difference can be observed by looking at the results. While conceptual models are comprehensible, reproducible, and explicit knowledge representations, AI techniques are capable of efficiently deriving an output from a given input while acting as a black box. AI solutions often lack comprehensiveness and reproducibility. Even the developers of AI systems can’t explain why a certain output is derived. In the Conceptual Modeling meets Artificial Intelligence (CMAI) workshop, we are interested in tackling the intersection of the two, thus far, mostly isolated approached disciplines of CM and AI. The workshop embraces the assumption, that manifold mutual benefits can be realized by i) investigating what Conceptual Modeling (CM) can contribute to AI, and ii) the other way around, what Artificial Intelligence (AI) can contribute to CM.

\end{abstract}

\section{Introduction}
Artificial Intelligence (AI) applications are conquering more and more domains in recent years with the increasing availability of vast amounts of data. The now dominant paradigm of data-driven AI employs big data to build intelligent applications and support fact-based decision making. The focus of data-driven AI is on learning (domain) models and keeping those models up-to-date by using statistical methods and machine learning (ML) over big data -- in contrast to the manual modeling approach prevalent in traditional, knowledge-based AI. While data-driven AI has led to signiﬁcant breakthroughs, it also comes with a number of disadvantages. First, models generated by AI often cannot be inspected and comprehended by a human being, thus lacking explainability and establishing challenges toward the utilization of such approaches in enterprises~\cite{Buxmann21-AI}. Furthermore, integration of pre-existing domain knowledge into learned models – prior to or after learning – is difficult.

In contrast to AI, conceptual modeling (CM) is human-driven. Humans are in charge of applying abstraction and creating a simplified representation of a system under study for a specific purpose. Consequently, conceptual models are comprehensible and foster understanding of existing, and the design of new systems. Models enable visual analysis, easy comprehension, and formalized representation by using a commonly agreed upon modeling language. Some of these positive attributes are, however, mitigated when models get very large (scalability) or the complexity of the system under study is still too high.

Only recently, first attempts aimed at combining the strengths of the two worlds while mitigating some of the mutual weaknesses. In this vein, the International Workshop on Conceptual Modeling Meets Artificial Intelligence (CMAI) was installed at ER after its initial inception at the German-speaking Conceptual Modeling conference 'Modellierung'~\cite{Reimer.20} in 2020. 

\section{Conceptual Modeling and Artificial Intelligence Research Framework}
First works like~\cite{bork2020ER} apply AI, in particular genetic algorithms and heuristic search, to automatically partition overarching data models into smaller, more comprehensible modules. Even fewer works attempt to show possibilities of improving AI-based systems by using conceptual modeling~\cite{lukyanenko2019using}.

\begin{figure}[t]
    \vspace{-.5cm}
    \centering
    \includegraphics[width=.66\linewidth]{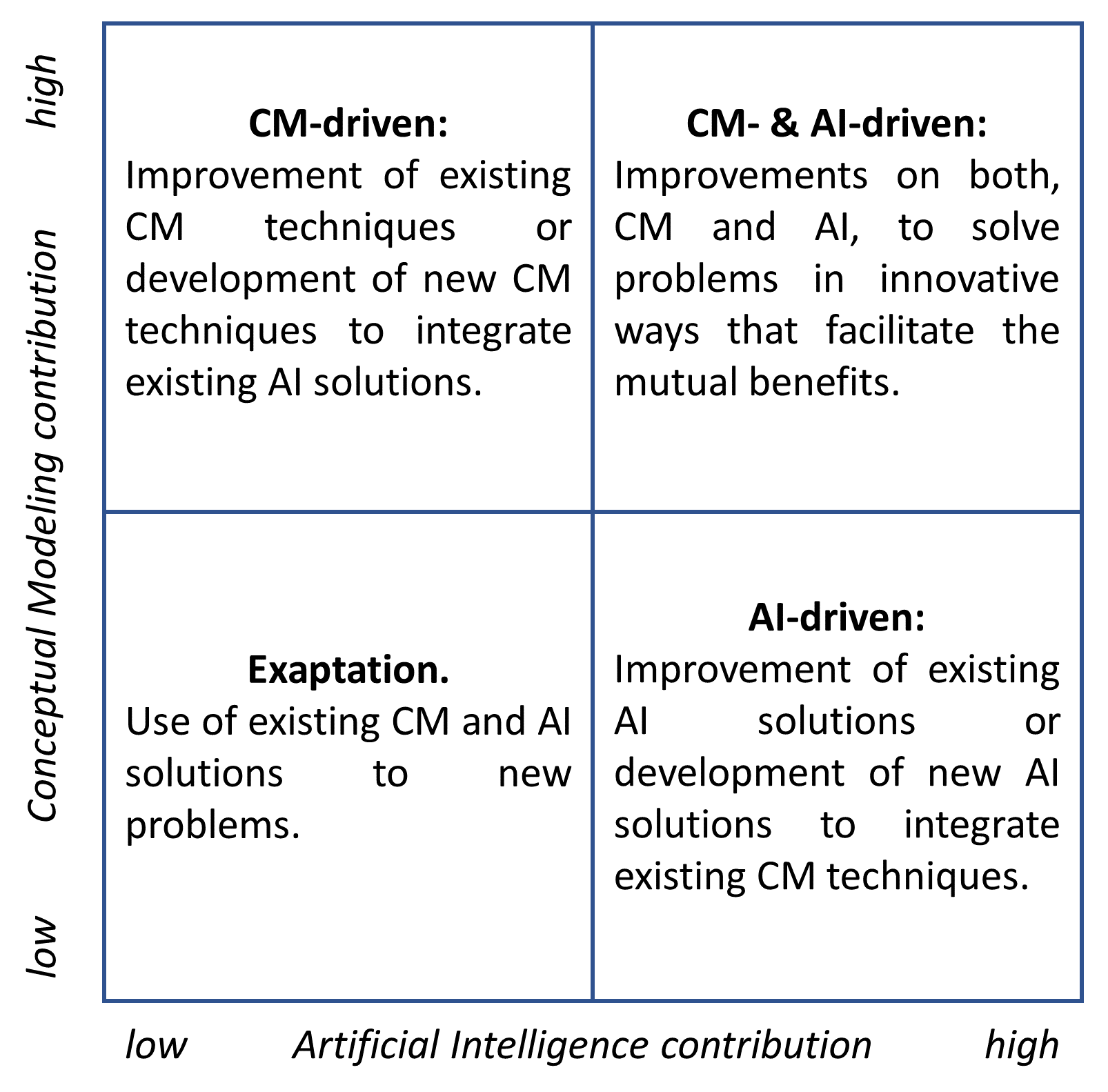}
    \caption{Conceptual Modeling and Artificial Intelligence Research Framework.}
    \vspace{-.1cm}
    \label{fig:cmai-contrib-framework}
    \vspace{-.35cm}
\end{figure}

In the following, we introduce a research framework for positioning and presenting conceptual modeling and artificial intelligence research. The framework, visualized in Fig.~\ref{fig:cmai-contrib-framework} positions the two research fields of conceptual modeling and artificial intelligence orthogonal to one another, thereby shaping four concrete CMAI research categories.

\begin{description}
\item[Exaptation CMAI.] Following the definition of exaptation research in the wider sense of Design Science Research given by Gregor and Hever~\cite{gregor2013positioning}, we define exaptation research in conceptual modeling and artificial intelligence as one that combines existing solutions from both fields to target a specific problem.

\item[CM-driven CMAI.] In this category of CMAI research fall all papers, that aim to improve existing or develop new conceptual modeling techniques to be combined with existing AI techniques to target a specific problem. Such research primarily contributes to the conceptual modeling research community.

\item[AI-driven CMAI.] In this category fall all papers, that aim to improve existing or develop new AI techniques to be combined with existing conceptual modeling techniques to target a specific problem. Such research primarily contributes to the artificial intelligence research community.

\item[CM- \& AI-driven CMAI.] In this category falls all CMAI research that makes contributions on both scientific fields, i.e., research that improves existing or develops new AI and CM techniques to target a specific problem.

\end{description}

\section{CMAI'2021 Program}
In 2021, the CMAI workshop was having its third edition, co-located with the ER'2021 conference. Due to a low number of submissions, CMAI'2021 was organized as a discussion-oriented workshop with invited, recent, high-quality papers fitting the scope. In the following, we will report on the program of the 3rd International Workshop on Conceptual Modeling Meets Artificial Intelligence (CMAI'2021). The papers composing the program will be briefly introduced and mapped to the CMAI research contribution framework (see Fig.~\ref{fig:cmai-contrib-framework}). 

Wolfgang Maass and Veda C. Storey outline in their paper entitled 'Why Should Machine Learning Require Conceptual Models?' the need for conceptual modeling contributions to machine learning and propose iterative cycles of a conceptualization-data-modeling (CDM) model. This paper fits best into the CM-driven category.

The paper entitled 'Conceptual Models for ML: Reflections and Guidelines' authored by Arturo Castellanos, Alfred Castillo, Monica Chiarini Tremblay, Roman Lukyanenko, Jeffrey Parsons and Veda C. Storey proposes the use of conceptual models to improve machine learning by augmenting ML training data with domain knowledge. This paper fits best into the CM-driven category.

Soroosh Nalchigar and Eric Yu contributed a paper about 'Using Conceptual Modeling to Drive Machine
Learning Solutions Development - A Case Report on applying GR4ML'. In this paper, the authors propose GR4ML, a conceptual modeling framework for requirements elicitation, design, and development of machine learning solutions. This paper fits best into the CM-driven category.

The paper entitled 'Searching for Models with Hybrid AI Techniques' by Martin Eisenberg, Hans-Peter Pichler, Antonio Garmendia and Manuel Wimmer reports on the MOMoT framework -- Marrying Optimization and Model Transformations which formulates the quest of finding the best models as an optimization problem with an emphasis on recent developments toward comparing the performance of different reinforcement learning techniques. This paper fits best into the CM\&AI-driven category as the comparison of the different AI techniques also contributes to the AI community.

Peter Fettke in his paper 'Conceptual Modelling and Artificial Intelligence - Overview and research challenges from the perspective of predictive business process management' overviews possible applications of AI for conceptual modeling and on conceptual modeling for AI. An example case refers to the predictive business process management. Due to this focus this extended abstract fits best into the AI-driven category.

\section{Concluding Remarks}
We want to thank all colleagues who were immediately willing to support the CMAI workshop with their valuable contributions. We believe this new format is equally valuable as conventional, publication-oriented workshops, in establishing a community and sharing the most recent and most interesting research results. Based on the experience of CMAI'2021, we will consider moving back to a publication-oriented workshop or even to establish a hybrid mode for the 2022 installment. In such a hybrid mode, recently published, high quality papers can be presented together with peer-reviewed direct CMAI workshop papers.

\bibliographystyle{splncs04}
\bibliography{cmai2021}

\end{document}